\documentclass[runningheads]{llncs}

\usepackage{eccv}

\usepackage{eccvabbrv}

\usepackage{graphicx}
\usepackage{booktabs}

\usepackage[accsupp]{axessibility}  

\usepackage{hyperref}

\usepackage{orcidlink}

\usepackage{multirow}
\usepackage{amssymb}
\usepackage{xcolor}
\definecolor{first}{RGB}{252, 191, 188}  
\definecolor{second}{RGB}{255, 237, 184} 
\definecolor{third}{RGB}{218, 232, 201}

\begin{document}

\title{TopoGS: Planar Reconstruction via Topology-aware 3D Gaussian Splatting} 

\titlerunning{TopoGS}

\author{Shanshan Pan\inst{1}\orcidlink{0009-0008-8259-5234} \and
Jiale Chen\inst{1}\orcidlink{0009-0005-5640-1836} \and
Yilin Liu\inst{2}\orcidlink{0000-0001-7336-1956} \and
Hui Huang\inst{1}\thanks{Corresponding author}\orcidlink{0000-0003-3212-0544}}

\authorrunning{S. Pan, J. Chen, Y. Liu, and H. Huang}

\institute{Guangdong Provincial Key Laboratory of Visual Media and Multidimensional Intelligence, CSSE, Shenzhen University, China \and
Autodesk Research, UK}

\maketitle

\begin{abstract}
    Extracting structured, parametric 3D representations from raw images remains a fundamental challenge in computer vision and graphics. 
    While recent advancements in the 3D Gaussian Splatting (3DGS) pipeline integrate planar primitives to yield compact and editable geometry, these approaches typically treat planes as isolated, discrete sets. 
    This lack of topological connectivity hinders robust geometric reasoning, leading to fragmented reconstructions and misaligned boundaries that fall short of the precision for rigorous spatial analysis and professional design workflows. 
    To address this, we introduce TopoGS, the first 3DGS framework to explicitly integrate both planar and topological constraints for coherent 3D reconstruction. 
    Specifically, we extract global 2D topological relationships from multi-view image segmentations and anchor Gaussian primitives to these structural elements. 
    This formulation enables the joint optimization of plane parameters, rendering fidelity, and topological adjacency. 
    By enforcing strict multi-view consistency alongside these topological constraints, our method significantly mitigates geometric misalignments and produces connected, structured 3D models. 
    Extensive evaluations on the ScanNet++ dataset demonstrate that TopoGS achieves state-of-the-art performance, providing a highly robust solution for generating accurate, topologically sound, and visually faithful scene representations.
\end{abstract}

\section{Introduction}
\label{sec:intro}

\begin{figure}
\centering
\includegraphics[width=\textwidth]{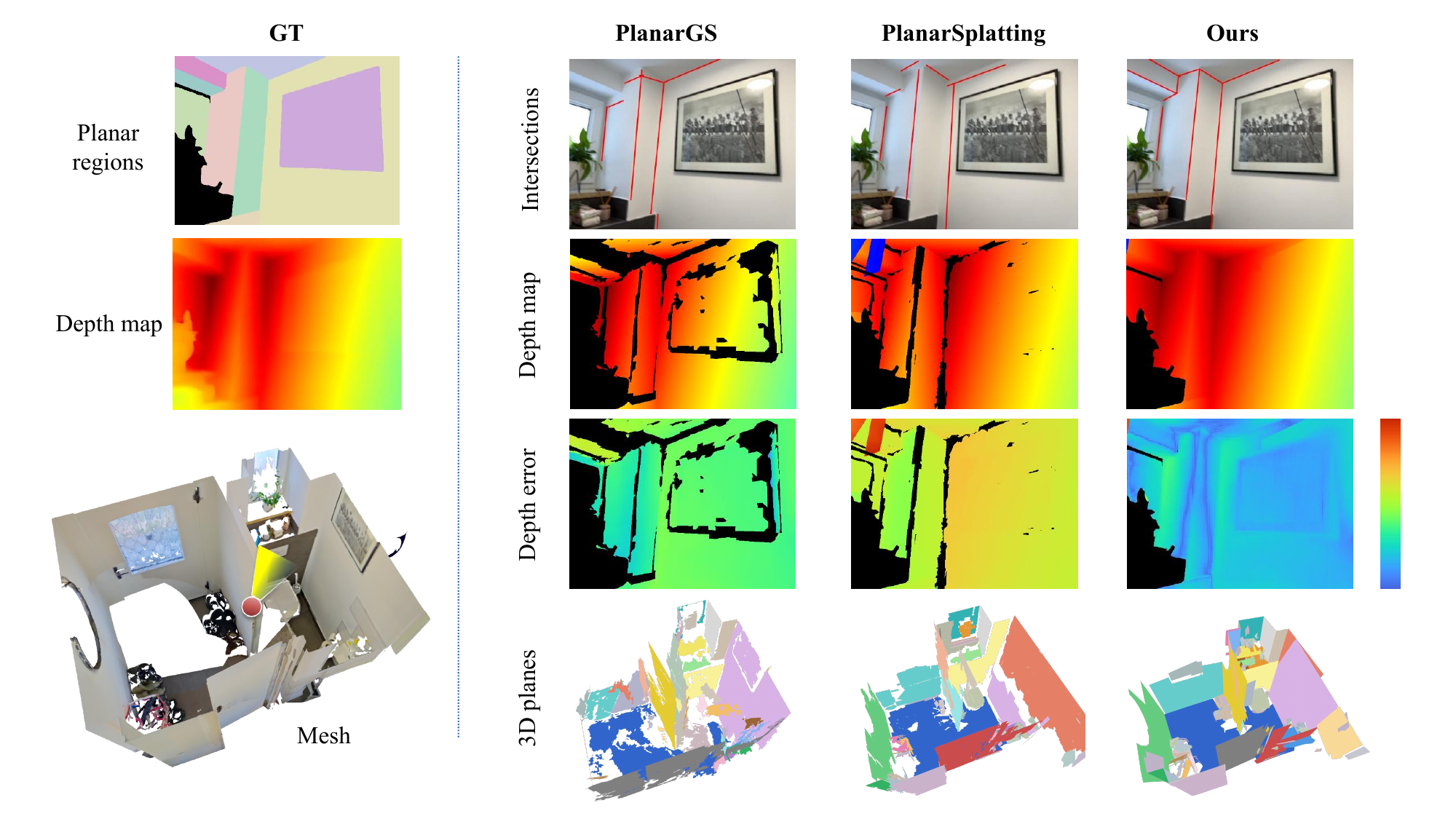}
\caption{\textbf{Motivation of TopoGS}. Existing methods show noticeable artifacts in plane intersection projections and depth error maps. In contrast, our TopoGS recovers precise plane boundaries and lower depth errors. This demonstrates that introducing topology constraints acts as a crucial regularizer, providing vital guidance for refining geometric accuracy and effectively suppressing noise in complex scenes. (In the error maps, blue indicates lower error while red signifies higher error.)}
\label{fig:motivation}
\vspace{-0.5cm}
\end{figure}

The creation of high-fidelity, editable 3D representations from raw multi-view images is a cornerstone problem in computer vision and computer graphics. For man-made environments and architectural spaces, the ideal abstraction is a structured planar model, a mathematically compact and interconnected representation explicitly defined by a set of parametric planes, alongside their intersecting lines and corner vertices. However, extracting this pristine structural information directly from unstructured image collections remains exceptionally challenging. Traditional reconstruction pipelines typically recover scene geometry as unorganized point sets or isolated surface fragments, treating scene elements as discrete entities. Consequently, the critical topological connectivity, denoting the explicit awareness of how these planes intersect, abut, or share boundaries, is frequently lost in the reconstruction process. This fundamental lack of structural awareness hinders robust geometric reasoning, inevitably producing fragmented models and misaligned boundaries that fall short of the precision required for downstream applications such as animation production, robotics, physical simulation, 3D printing, and AR/VR.

Historically, efforts to bridge this structural gap and recover coherent planar models have predominantly relied on a decoupled, two-stage Multi-View Stereo (MVS) pipeline \cite{luo2016efficient}\cite{yao2018mvsnet}\cite{sayed2022simplerecon}\cite{bauchet2020kinetic}\cite{he2024windpoly}. In the first stage, these methods process multi-view source images to obtain a dense intermediate point cloud. In the second stage, they detect primitive planes from this point cloud and employ complex spatial partitioning to construct a cohesive structural representation. However, the efficacy of this algorithm is fundamentally bottlenecked by the quality of the intermediate geometry. This vulnerability is especially pronounced in indoor scenes, where textureless walls, reflective surfaces, and varying illumination frequently cause both traditional and learning-based MVS techniques to produce noisy, outlier-ridden, or incomplete point clouds. Consequently, the subsequent plane detection becomes highly inaccurate, leading to the omission of critical structural details. Because the final structuring phase relies entirely on this flawed intermediate representation, it fails to explicitly leverage the rich photometric cues and multi-view consistency present in the raw images, inevitably accumulating severe errors.

To bypass the limitations of intermediate point clouds, recent frameworks increasingly optimize scene representations directly from multi-view images
\cite{xie2022planarrecon}\cite{ye2025neuralplane}\cite{zanjani2025planar}\cite{chen2024pgsr}\cite{jin2025planargs}\cite{ruan2025indoorgs}. 
Notably, 3D Gaussian Splatting (3DGS)\cite{kerbl20233dgs} has emerged as a highly efficient rendering paradigm. By integrating 3DGS with foundation vision models, recent approaches \cite{tan2025planarsplatting}\cite{gan2025gsplane}\cite{zanjani2025planar} leverage data-driven monocular priors (e.g., depth, normals, and semantics) to directly recover planar primitives during the optimization. While these 2D cues significantly improve geometric perception in challenging regions, current 3DGS-based methods still fail to achieve the requisite structural awareness. By continuing to treat recovered planes as isolated, independent entities, they remain completely devoid of topological connectivity. However, physical environments are governed by strict structural dependencies, such as the intersecting corners of adjacent walls or the contact boundaries between furniture and floors. We contend that explicitly integrating these topological relationships as vital geometric constraints, rather than relegating them to optional post-processing, can effectively guide the 3DGS optimization landscape, enforce global structural coherence, and dramatically enhance robustness against noise (see Fig.~\ref{fig:motivation}).

To this end, we introduce TopoGS, a novel 3DGS framework designed to reconstruct explicitly connected, structured planar models. Our pipeline begins at the image level, extracting 2D region segmentations and constructing their corresponding adjacency graphs directly from the multi-view source images. However, naively lifting these 2D relationships into 3D is fraught with error. 
Because of perspective occlusions and depth discontinuities, regions that appear adjacent or overlapping in a 2D projection are often mistakenly identified as physically intersecting in 3D. To systematically distill these ambiguous 2D observations into verified 3D topology, we anchor our Gaussian primitives to the segmented regions. By embedding these anchored primitives within the 3DGS optimization loop, we enforce a tri-consistency constraint, integrating photometric, geometric, and topological facets, to jointly refine appearance, scene geometry and plane parameters. This optimization acts as a natural filter, disambiguating genuine physical connections from spurious, view-dependent adjacencies. Finally, we actively enforce these verified 3D adjacencies to assemble the optimized primitives into a globally consistent, structured 3D representation.

Our main contributions are summarized as follows:
\begin{itemize}
    \item We introduce a novel 3DGS-based framework that integrates geometry and topology constraints to recover consistent plane geometry and topology. 
    \item We propose a method to extract connected structured planar models by leveraging optimized, reliable 3D adjacent relationships.
    \item Our approach achieves state-of-the-art performance on high-fidelity planar reconstruction tasks within the ScanNet++ dataset.
\end{itemize}

\section{Related Work}

\subsection{Two-Stage Plane Reconstruction}
Traditional approaches for recovering structured planar models typically rely on a decoupled, two-stage paradigm: i) front-end geometry reconstruction to obtain a dense representation (e.g., a point cloud or mesh), followed by ii) back-end plane detection and spatial partitioning to assemble the final structured model.

The primary target of the first stage is to extract raw, dense scene geometry from multi-view source images.
Early methods relied on Multi-View Stereo (MVS) \cite{bleyer2011patchmatch}\cite{luo2016efficient}\cite{yao2018mvsnet}\cite{sayed2022simplerecon}\cite{stier2023finerecon}, which achieves high precision in textured areas but often fails in featureless indoor environments. To mitigate this, neural representations like NeRF and 3DGS incorporate geometric constraints, such as normal regularization and planar priors to enhance surface quality. For instance, PGSR\cite{chen2024pgsr} introduces multi-view consistency, while IndoorGS\cite{ruan2025indoorgs} incorporates line and plane priors to bolster indoor reconstruction capabilities. PlanarGS\cite{jin2025planargs} further leverages language-prompt planar priors to provide more reliable geometric guidance. However, these per-scene optimization methods entail heavy computational overhead. Consequently, recent feed-forward approaches \cite{yang2024depth}\cite{yang2024depth2}\cite{lin2025depth3}\cite{wang2025vggt}\cite{wang2025pi} leverage Transformer-based attention for cross-view alignment, providing robust geometric priors without iterative optimization.

In the second stage, the target shifts to extracting and structuring semantic primitives from this intermediate geometry. This process typically begins with plane detection via point clustering
(e.g., RANSAC\cite{schnabel2007efficient} or region growing\cite{verdie2015lod}) followed by spatial partitioning (e.g., kinetic or binary partitioning \cite{bauchet2020kinetic}\cite{chen2022reconstructing}\cite{he2024windpoly}). Despite improvements in front-end geometry, this two-stage workflow suffers from unidirectional error propagation: front-end noise directly misleads back-end plane fitting. Moreover, the plane assembly process cannot back-propagate to refine the front-end geometry, and the rich information in raw images is largely ignored during the second stage, limiting the fidelity of the final planar model.

\subsection{Plane Reconstruction from Images}
Distinct from two-stage methods, end-to-end planar reconstruction aims to generate plane geometry directly from multiple views. Early research \cite{xie2022planarrecon} \cite{sun2021neuralrecon} focused on using deep networks to predict planar parameters in voxel space or at the pixel level. With the rise of neural rendering, focus has shifted toward integrating planar primitives into the rendering pipelines of NeRF or 3DGS. For example, NeuralPlane\cite{ye2025neuralplane} jointly optimizes plane geometry and NeRF representations. PlanarSplatting \cite{tan2025planarsplatting} mimics the alpha-blending capability of 3DGS by proposing plane primitives to replace traditional Gaussian primitives. Similarly, PGS \cite{zanjani2025planar} leverages 3D Gaussian primitives for scene modeling but introduces a hierarchical, tree-structured Gaussian Mixture Model (GMM) to group adjacent Gaussian primitives into cohesive planar instances, thereby mitigating the issue of plane over-segmentation. Moreover, GSPlane\cite{gan2025gsplane} and GSFlats \cite{taktasheva20253d} represent scenes as a hybrid of planes and triangles, categorizing Gaussian primitives into planar and non-planar types to adaptively model scenes with both flat and curved surfaces.

While these end-to-end methods successfully enhance local surface smoothness through geometric regularization, they fundamentally treat each plane as an isolated geometric entity. Lacking a global topological graph to explicitly constrain spatial intersections, these approaches often produce misaligned boundaries and fragmented planar shapes, especially under severe occlusion or high noise. Consequently, collaboratively optimizing low-level geometric priors with high-level topological constraints within a unified neural rendering framework remains a critical, unresolved challenge.

\section{Method}
\label{sec:method}

\begin{figure}[t]
\centering
\includegraphics[width=\textwidth]{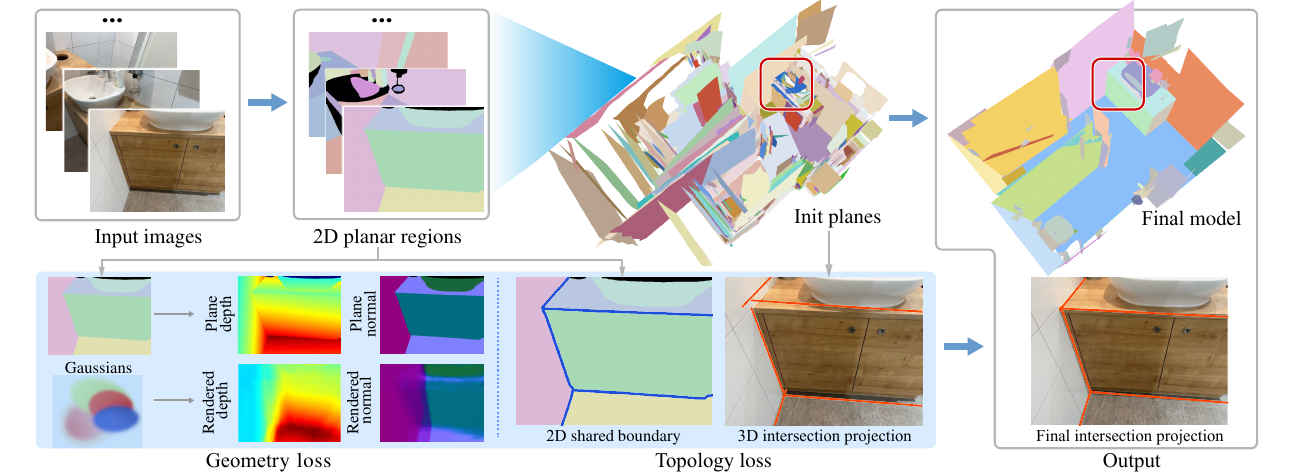}
\caption{\textbf{Overview of TopoGS.} (1) \textbf{Initialization}: 2D plane regions are extracted from input images and back-projected into 3D space. (2) \textbf{Joint optimization}: Plane parameters are refined within the 3DGS-based framework, supervised by our proposed geometry and topology losses. (3) \textbf{Structural assembly}: The optimized primitives are integrated into a structured planar model, characterized by topologically consistent and accurate intersection boundaries between adjacent planes.}
\label{fig:overview}
\vspace{-0.3cm}
\end{figure}

Given a set of images with camera poses, our goal is to recover the underlying scene structure represented by a connected plane model. We propose TopoGS, a 3DGS-based framework designed to reconstruct a structured model consisting of a set of vertices $\mathcal{V} \subset \mathbb{R}^3$, a set of intersection edges $\mathcal{E} = \{(v_i, v_j) \mid v_i, v_j \in \mathcal{V}\}$, and a set of bounded planes $\mathcal{P} = \{(\mathbf{\pi}_k, E_k) \mid E_k \subset \mathcal{E}\}$, where $\pi_k = (\mathbf{n}_k, d_k) \in \mathbb{R}^4$ represents the plane parameters and $E_k$ denotes the index of the bounding edges. 
As illustrated in Fig.~\ref{fig:overview}, our method comprises three primary stages: (1) Initialization, which extracts plane primitives and their corresponding image-space adjacencies; (2) Joint optimization, which simultaneously refines appearance, scene geometry and plane parameters; and (3) Structural assembly, which converts the optimized primitives into a globally consistent, structured planar model by filtering out false-positive adjacencies.

\subsection{Geometry \& Topology Initialization}
Given a set of images $\{\mathbf{I}_k\}$ and camera poses $\{\mathbf{T}_k\}$, this section aims to recover: i) for each image $\mathbf{I}_k$, a set of 2D planar regions $\{\mathcal{S}_k\}$ and their corresponding plane parameters $\{\pi_i\}$; and ii) a 2D adjacency matrix $\mathcal{A}_{2D}$ that encodes the topological relationships between these regions. These initializations serve as the geometric and topological constraints for subsequent optimization.
Following the practices of PlanarSplatting\cite{tan2025planarsplatting}, we leverage a pre-trained Metric3D model\cite{hu2024metric3d} to predict single-view depth maps $\mathbf{D}_{pred}$ and normal maps $\mathbf{N}_{pred}$ to assist in the initialization.

\paragraph{Hierarchical Segmentation.} 
To ensure that segmentation boundaries align with both semantic logic and geometric consistency, we adopt a two-stage strategy. 
First, we use the Segment Anything Model (SAM)\cite{kirillov2023segment} to obtain initial semantic masks $\mathcal{M}_{sem}$ for each image.
These masks provide high-fidelity structural boundaries, effectively delineating critical geometric features such as wall-to-wall junctions. 
Since a single semantic entity may consist of multiple geometric planes, we further perform Mean-shift clustering\cite{meanshift} within each masked region based on the predicted normal map $\mathbf{N}_{pred}$. 
This subdivides the masks into a set of planar regions $\{\mathcal{S}_k\}$ for each image, where we preserves the precise edges of SAM while capturing geometric transitions.

\paragraph{World-Space Plane Fitting.} To lift 2D planar regions into 3D space, we fit plane parameters $\pi_i$ for each planar region $s_i \in \mathcal{S}$ using the predicted depth maps $\mathbf{D}_{pred}$. 
Specifically, for a pixel $p \in s_i$ with homogeneous coordinates $\mathbf{u}_p$, its 3D position $\mathbf{X}_p$ in the world coordinate system is back-projected via:
\begin{equation}
\mathbf{X}_p = \mathbf{R}_k \left( D_{\text{pred}}(p) \cdot \mathbf{K}^{-1} \mathbf{u}_p \right) + \mathbf{t}_k,
\label{eq:pixel-to-world}
\end{equation}
where $\mathbf{T}_k = [\mathbf{R}_k | \mathbf{t}_k]$ denotes the camera pose (Camera-to-world matrix) and $\mathbf{K}$ is the intrinsic matrix of the camera. We then employ the RANSAC method\cite{schnabel2007efficient} to estimate the plane parameters from these points.

\paragraph{Adjacency Construction.} 
The initial adjacency graph $\mathcal{A}_{2D}$ inevitably contains spurious connections, such as curved boundaries or segmentation noise. 
These invalid candidates, which inherently violate the linearity assumption of planar intersections, can be filtered out first to support subsequent optimization. 
For a pair of adjacent regions $(s_i, s_j)$, we extract their shared boundary pixels $\mathcal{B}_{ij}$ and try to fit a straight line to these pixels.
Adjacencies with significant non-linear residuals or insufficient contact length are discarded as they fail to represent valid 3D intersection lines. 

The final output includes 2D planar regions $\{\mathcal{S}_k\}$ and a refined 2D adjacency graph $\mathcal{A}_{2D}$. 
Unlike NeuralPlane\cite{ye2025neuralplane}, our initialization offers: (1) Topological completeness: better 2D adjacency identification via complete segmentation; (2) Boundary accuracy: preservation of sharp SAM edges within the normal-based subdivision; and (3) Robustness: enhanced plane recovery in textureless regions by utilizing predicted depth instead of noisy SFM points.

\subsection{Joint Optimization}
Directly merging initial plane primitives often fails to yield a coherent scene representation, as it is highly susceptible to multi-view inconsistencies (see Fig.~\ref{fig:overview}, "Init planes"). 
To address this, we propose an explicit optimization framework built upon 3D Gaussian Splatting (3DGS) to jointly refine both the geometry and topology of the planes.

\paragraph{Plane-Anchored Gaussian Initialization.} 
To facilitate differentiable rendering, we anchor Gaussian primitives to 2D planar regions $\{\mathcal{S}_k\}$. For each view $I_k$, pixels are sampled on a uniform grid with a stride of $(H/4, W/4)$. For each sampled pixel $p$, we recover its 3D position $\mathbf{X}_p$ via back-projection using Eq.~\ref{eq:pixel-to-world}. These 3D coordinates serve as the initial means for the Gaussians, while other attributes are initialized following standard 3DGS protocols.

\paragraph{Loss Functions.} We optimize the scene via differentiable Gaussian splatting to synthesize the rendered color $\hat{C}$, depth $\hat{D}$, and normal $\hat{\mathbf{N}}$. In addition to the standard photometric loss $\mathcal{L}_{rgb}$ and the scale loss $\mathcal{L}_{scale}$ proposed in PGSR~\cite{chen2024pgsr}, we introduce two additional key constraints to regularize the optimization.

(1) Geometry Loss ($\mathcal{L}_{geo}$): To ensure that the 3DGS-rendered geometry remains consistent with the parametric planes, we minimize the discrepancy between the rendered depth $\hat{D}(p)$ and the computed plane depth $D_{\pi}(p)$, similarly to that between the rendered normal $\hat{\mathbf{N}}(p)$ and the plane normal $\mathbf{N}_{\pi}(p)$. 
We define the plane depth $D_{\pi}(p)$ as the orthogonal distance (Z-depth) from the camera center to the intersection point of the ray and the plane. 
The geometry loss is then formulated as:
\begin{equation}
\mathcal{L}_{geo} = \sum_{p \in \mathcal{I}_k} \lambda_1 \left( | \hat{D}(p) - D_{\pi}(p) |_1 \right)+\lambda_2 \left( | \hat{\mathbf{N}}(p) - \mathbf{N}_{\pi}(p) |_1 \right),
\end{equation}
where $\lambda_1$ and $\lambda_2$ are weighting factors. 

(2) Topology Loss ($\mathcal{L}_{topo}$): 
To enforce physical closure and structural consistency between two adjacent planes $(s_i, s_j)$, we introduce a topology loss that constrains the projected 3D intersection lines to align with the observed shared boundaries $\mathcal{B}_{ij}$ in the 2D pixel space. 
For any pair of adjacent regions $(s_i, s_j) \in \mathcal{A}_{2D}$ with corresponding planes $\pi_i = (\mathbf{n}_i, d_i)$ and $\pi_j = (\mathbf{n}_j, d_j)$, their 3D intersection line $\mathbf{L}_{ij}$ is analytically determined by the system of linear equations:
\begin{equation}
\mathbf{n}_i \cdot \mathbf{X} + d_i = 0, \mathbf{n}_j \cdot \mathbf{X} + d_j = 0.
\end{equation}
We represent $\mathbf{L}_{ij}$ by two 3D points $\{X_0, X_1\}$ and project them onto the image using the camera intrinsic matrix $\mathbf{K}$ and extrinsic parameters $[\mathbf{R}|\mathbf{t}]$ to obtain the 2D projected line $\mathbf{l}_{ij}$ defined by points $\{x_0, x_1\}$.
The topology loss is defined as the weighted sum of the mean distances across all adjacent pairs:
\begin{equation}
\mathcal{L}_{topo} = \sum_{(s_i, s_j) \in \mathcal{A}_{2D}} w_{ij} \cdot \mu_{ij},
\end{equation}
where $\mu_{ij} = \frac{1}{|\mathcal{B}_{ij}|} \sum_{p \in \mathcal{B}_{ij}} d(p, \mathbf{l}_{ij})$, representing the mean Euclidean distance from all pixels $p$ within $\mathcal{B}_{ij}$ to the ideal line $\mathbf{l}_{ij}$ defined by the vertex pair $(x_0, x_1)$.

Notably, 2D adjacencies $\mathcal{A}_{2D}$ extracted from segmentations are often "spurious", as they may be caused by perspective occlusions or depth discontinuities rather than true physical contact in 3D. To prevent these misleading 2D observations from degrading the 3D geometry, we introduce an uncertainty-aware weighting mechanism $w_{ij}$:
\begin{equation}
w_{ij} = \exp\left( -\frac{\mu_{ij} + \sigma^2_{ij}}{\tau} \right),
\end{equation}
where $\sigma^2_{ij}$ is the variance of the distances $d(p, \mathbf{l}_{ij})$ for pixels in $\mathcal{B}_{ij}$, and $\tau$ is a temperature hyperparameter (set to 50.0). The intuition behind this design is two-fold: i) filtering spurious adjacencies: if two regions are adjacent in 2D but far apart in 3D (high $\mu_{ij}$), or if their shared boundary is geometrically inconsistent with a single straight line (high $\sigma^2_{ij}$), the weight $w_{ij}$ decays exponentially and ii) self-driven disambiguation: this allows the optimization loop to act as a natural filter. It prioritizes the refinement of "genuine" physical connections where the 2D pixels already show a strong consensus with the 3D plane intersection, while effectively ignoring view-dependent occlusions that cannot be reconciled with the 3D topology.

\paragraph{Optimization Schedule.} 
We employ a staged strategy to move from coarse geometry to refined topology:
\begin{itemize}
    \item Stage 1: Coarse optimization (0–4k iterations): Optimize $\mathcal{L}_{rgb}$, $\mathcal{L}_{scale}$, and $\mathcal{L}_{geo}$ to establish a stable geometric foundation.
    \item Stage 2: Fine optimization (4k–6k iterations): Introduce $\mathcal{L}_{topo}$ to "snap" adjacent planes together and resolve structural ambiguities.
    \item Adaptive Plane Control: Every 2,000 iterations, we perform two plane operations: merging (combining planes with similar normals/offsets and overlaps) and pruning (removing low-confidence planes) to maintain a compact representation.
\end{itemize} 
The objective function transitions as follows:
\begin{equation}
\mathcal{L} = 
\begin{cases} 
\lambda_{1}\mathcal{L}_{rgb} + \lambda_{2}\mathcal{L}_{scale} + \lambda_{3}\mathcal{L}_{geo} & t < 4000, \\
\mathcal{L}_{Stage1} + \lambda_{4}\mathcal{L}_{topo} & t \ge 4000. 
\end{cases}
\end{equation}

\subsection{Structural Assembly}
The final stage of our pipeline assembles the optimized parametric planes and their verified topologies into a coherent, structured 3D model. This process transitions from discrete planes to a connected mesh-like representation.

\textbf{Adjacency Verification.} We first refine the initial 2D adjacency graph $\mathcal{A}_{2D}$ to extract a set of verified 3D adjacencies $\mathcal{A}_{verified}$. Benefiting from the joint optimization in Stage 2, genuine physical connections exhibit high geometric consensus with the 2D shared boundaries $\mathcal{B}_{ij}$. This allows us to apply a strict distance threshold to $\mu_{ij}$ to effectively prune spurious adjacencies and retain only true-positive physical intersections.

\textbf{Intersection Geometry Extraction.} For each verified pair in $\mathcal{A}_{verified}$, the 3D intersection line $\mathbf{L}_{ij}$ is analytically derived from the optimized parameters of planes $\pi_i$ and $\pi_j$. To convert these infinite lines into finite edges, we identify shared corners among the 2D boundary segments $\mathcal{B}_{ij}$: if two segments share a common corner, their corresponding 3D lines are deemed adjacent. The 3D corner vertices are then computed as the intersection points of these adjacent lines and used as endpoints to truncate $\mathbf{L}_{ij}$ into bounded segments, forming the skeletal framework of the structured model.

\textbf{Multi-View Boundary Fusion and Refinement.} Since a single perspective often captures only a fragmented portion of a large planar surface, we aggregate the boundary segments of each plane primitive $i$ from all visible multi-view images into a unified global boundary. To ensure structural integrity, we use previously established 3D intersection lines and corner vertices to refine these fused boundaries. This constrained trimming process ensures that adjacent planes converge precisely at their shared geometric anchors, effectively eliminating gaps or overlaps. By combining the refined plane boundaries, intersection lines, and corner vertices, we assemble the primitives into a connected structured model.

\section{Experiments}

\subsection{Dataset, Metrics \& Baselines}
\paragraph{Datasets.} To validate the effectiveness of our proposed framework, we conduct extensive evaluations on ScanNet++ datasets\cite{yeshwanth2023scannet++}, a benchmark widely recognized\cite{ye2025neuralplane}\cite{tan2025planarsplatting}\cite{jin2025planargs} for its high-fidelity reconstructions and dense annotations. Specifically, we evaluate our method on 25 randomly selected scenes. Following the preprocessing protocols of PlaneRCNN \cite{liu2019planercnn}, we employ their provided scripts to derive 3D planar ground truth from the raw meshes.

\paragraph{Evaluation Metrics.} Following the evaluation framework established by AirPlanes\cite{watson2024airplanes}, we evaluate our method using four categories of metrics:
\begin{itemize}
  \item \textbf{Geometry:} We report Chamfer Distance (CD) and F-score ($5\text{cm}$ threshold) to measure the global geometric alignment with the GT dense mesh.
  \item \textbf{Planar quality:} We evaluate the reconstruction quality of Top-20 largest planes \cite{watson2024airplanes} from the ground truth and use the metrics including Planar Fidelity, Accuracy, and Chamfer.
  \item \textbf{Segmentation:} To evaluate 3D plane segmentation, we uniformly sample 200k points on the generated planes to address the disparity in representation density. These points are then used to calculate standard metrics, including Variation of Information (VOI), Rand Index (RI), and Segmentation Covering (SC).
  \item \textbf{Topology:} Since our approach explicitly recovers vertices and edges, we further assess the F-score for these entities, as well as for planes. We apply multiple thresholds (0.05, 0.1, 0.2, 0.5) for a thorough evaluation. 
\end{itemize}

\paragraph{Baselines.} We compare our approach against diverse state-of-the-art methods:
\begin{itemize}
  \item \textbf{Two-stage Methods:} These pipelines involve dense reconstruction followed by plane extraction. We select two representative dense reconstruction methods for comparison: (1) PlanarGS \cite{jin2025planargs}, a 3DGS-based approach incorporating planar priors, and (2) Depth-Anything-V3 \cite{lin2025depth3}, a high-performance feed-forward depth estimation model. For these baselines, we employ the RANSAC-based plane detection protocol provided by AirPlanes \cite{watson2024airplanes} to derive planar models from the generated meshes.
  \item \textbf{Direct Planar Reconstruction:} We compare against end-to-end methods that reconstruct planes directly, including NeuralPlane\cite{ye2025neuralplane} (NeRF-based), and PlanarSplatting\cite{tan2025planarsplatting} (3DGS-based). These methods do not require RANSAC-based post-processing. 
\end{itemize}
For all learning-based baselines, we utilize the officially released pre-trained weights to ensure an equitable comparison.

\subsection{Comparisons with baselines}
\begin{table}[t]
\centering
\newcommand{\first}[1]{\colorbox{first}{#1}}
\newcommand{\second}[1]{\colorbox{second}{#1}}
\newcommand{\third}[1]{\colorbox{third}{#1}}
\caption{\textbf{Quantitative evaluation on the ScanNet++ dataset.} We compare our proposed TopoGS against state-of-the-art two-stage pipelines (combined with AirPlanes' RANSAC extraction) and direct planar reconstruction methods. Top-3 results are highlighted as \first{first}, \second{second}, and \third{third}. Our method achieves leading performance in most planar and geometry metrics, demonstrating the effectiveness of the proposed tri-consistency optimization.}
\label{tab:results}
\resizebox{\linewidth}{!}{
\begin{tabular}{l|cc|ccc|ccc}
\hline
\multirow{2}{*}{Method} & \multicolumn{2}{c|}{Geometry} & \multicolumn{3}{c|}{Planar} & \multicolumn{3}{c}{Segmentation} \\ \cline{2-9} 
 & CD $\downarrow$ & F-score $\uparrow$ & Fidelity $\downarrow$ & Acc $\downarrow$ & CD $\downarrow$ & VOI $\downarrow$ & RI $\uparrow$ & SC $\uparrow$ \\ \hline
PlanarGS\cite{jin2025planargs} + Airplanes   &14.02 &23.77 &24.74 &19.25 &22.00 &3.400 &0.926 &0.379 \\
DA3\cite{lin2025depth3} + AirPlanes          & 8.49 &\first{74.40} &20.35 &12.85 &16.60 &3.041 &0.919 &0.431 \\ \hline
NeuralPlane\cite{ye2025neuralplane}         &\third{7.56} &53.83 &\third{14.50} &\third{11.83} &\third{13.17} & \third{2.872} & \third{0.944} & \second{0.464} \\
PlanarSplatting\cite{tan2025planarsplatting} &9.06 &53.83 &15.07 &14.06 &14.56 & \second{2.869} & 0.933 & \third{0.463}\\
Ours                                         & \first{6.01} & \second{70.18} & \first{12.34} & \first{9.96} & \first{11.15} & 2.933 & \first{0.946} & \third{0.463}\\ 
Ours (+ Assembly)                         & \second{6.09} & \third{69.80} & \second{12.96} & \second{10.91} & \second{11.94} & \first{2.867} & \first{0.946} & \first{0.472}\\ \hline
\end{tabular}
}
\end{table}

As summarized in Table \ref{tab:results}, our proposed TopoGS consistently outperforms both two-stage pipelines and direct planar reconstruction methods across the majority of evaluation metrics. Specifically, in terms of planar reconstruction, TopoGS achieves a significant lead with a Fidelity of 12.34 and an Accuracy (Acc) of 9.96, substantially surpassing the state-of-the-art direct method, NeuralPlane\cite{ye2025neuralplane}. This performance gain underscores the efficacy of our tri-consistency optimization in refining planar parameters.

\begin{figure}
    \centering
    \includegraphics[width=0.98\linewidth]{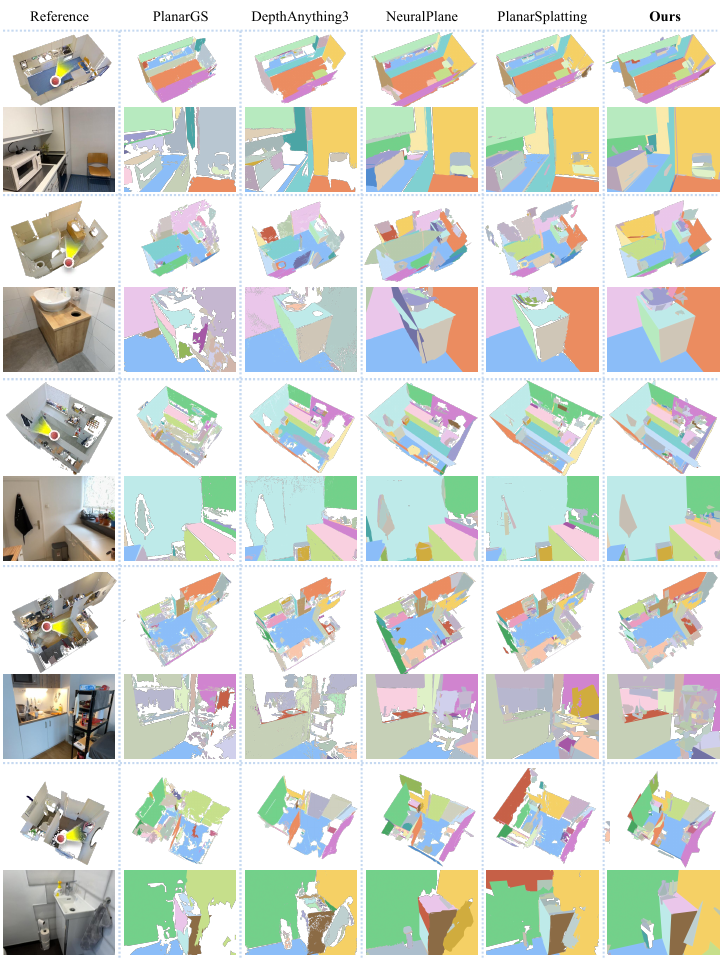}
    \caption{\textbf{Qualitative comparison on the ScanNet++ dataset.} We evaluate our method against state-of-the-art baselines across diverse environments, including kitchens, bathrooms, and living areas. Compared to existing two-stage and direct reconstruction pipelines, our framework achieves higher structural fidelity and global consistency. Our results exhibit fewer artifacts and more precise boundaries, particularly in challenging regions characterized by weak textures or complex occlusions.}
    \label{fig:baseline_results}
\end{figure}

While Depth-Anything-V3 \cite{lin2025depth3} reports a competitive F-score (74.40), attributable to its robust pre-trained depth priors and confidence-based pruning, it fails to recover precise planar structures, as evidenced by its considerably higher planar error metrics. We attribute this deficiency to two factors: first, the difficulty Depth-Anything-V3 faces in textureless or reflective regions, leading to incomplete reconstructions; second, the error propagation inherent in two-stage pipelines where depth estimation and plane extraction are decoupled.

Figure \ref{fig:baseline_results} provides a visual comparison between TopoGS and the baseline methods. As illustrated by the qualitative results, TopoGS consistently generates the most structured and visually coherent 3D models across diverse indoor scenes, ranging from cluttered kitchens to complex bathroom fixtures.

Despite the success of planar priors in recovering textureless areas (as demonstrated by NeuralPlane\cite{ye2025neuralplane}, PlanarSplatting \cite{tan2025planarsplatting}, and PlanarGS \cite{jin2025planargs}), these methods still struggle with inherent geometric inaccuracies on featureless walls and floors. This occurs because textureless regions introduce photometric ambiguity, settling the optimization into scale-deviant local optima that satisfy planarity and multi-view consistency (e.g., via normalized cross correlation (NCC)) despite being spatially inaccurate. In contrast, our topological constraints capitalize on the semantic understanding of SAM\cite{kirillov2023segment}. By utilizing SAM’s ability to delineate structural boundaries (such as wall-to-wall intersections) even in textureless regions, our model identifies geometric scaling discrepancies that would otherwise lead to misalignments between projected plane intersections and image segmentation boundaries. Consequently, enforcing these inter-plane topological constraints yields a marked improvement across all metrics.

Furthermore, in challenging scenarios with heavy occlusion, glass reflections, or high sensor noise, Depth-Anything-V3 and PlanarGS often produce inconsistent or noisy depth maps, leading to significant geometric holes. When integrated with AirPlanes\cite{watson2024airplanes} for plane extraction, these errors accumulate. Our framework maintains a tighter coupling with the source imagery, combined with topological constraints, effectively resolves ambiguities in these regions and restores a more complete visible range.

Finally, TopoGS excels at recovering sharp, unambiguous boundaries. Unlike baselines that treat the scene as a collection of discrete primitives, TopoGS generates models characterized by physically plausible connections and a "watertight" tendency between adjacent planes. This qualitative superiority reinforces our quantitative findings, demonstrating that explicit topological modeling is indispensable for achieving high-fidelity indoor scene reconstruction.

\subsection{Ablation Studies}
\begin{table}[t]
\centering
\caption{\textbf{Impact of tri-consistency constraints.} This ablation study demonstrates that while photometric consistency provides a baseline, the addition of geometry and topology losses is crucial for improving geometry quality. Our full pipeline significantly outperforms the ablated versions, validating the synergy between the proposed constraints.}
\label{tab:ablation}
\begin{tabular}{l|cc|ccc|ccc}
\hline
\multirow{2}{*}{Method} & \multicolumn{2}{c|}{Geometry} & \multicolumn{3}{c|}{Planar}  & \multicolumn{3}{c}{Topology (F-score)$\uparrow$} \\ \cline{2-9} 
 & CD $\downarrow$ & F-score $\uparrow$ & Fidelity $\downarrow$ & Acc $\downarrow$ & CD $\downarrow$ & V & E & P \\ \hline
w/o Topo      & 6.54 & 64.98 & 13.98 & 10.96 & 12.47 & 19.62 & 32.05 & 34.22 \\
w/o Geom      & 7.56 & 62.32 & 16.50 & 10.66 & 13.58 & 12.72 & 21.28 & 33.38 \\ 
Full loss     & \textbf{6.01} & \textbf{70.18} & \textbf{12.34} & \textbf{9.96} & \textbf{11.15} & \textbf{21.52} & \textbf{35.18} & \textbf{36.55} \\ \hline
\end{tabular}
\end{table}

\begin{figure}[!ht]
    \centering
    \includegraphics[width=\linewidth]{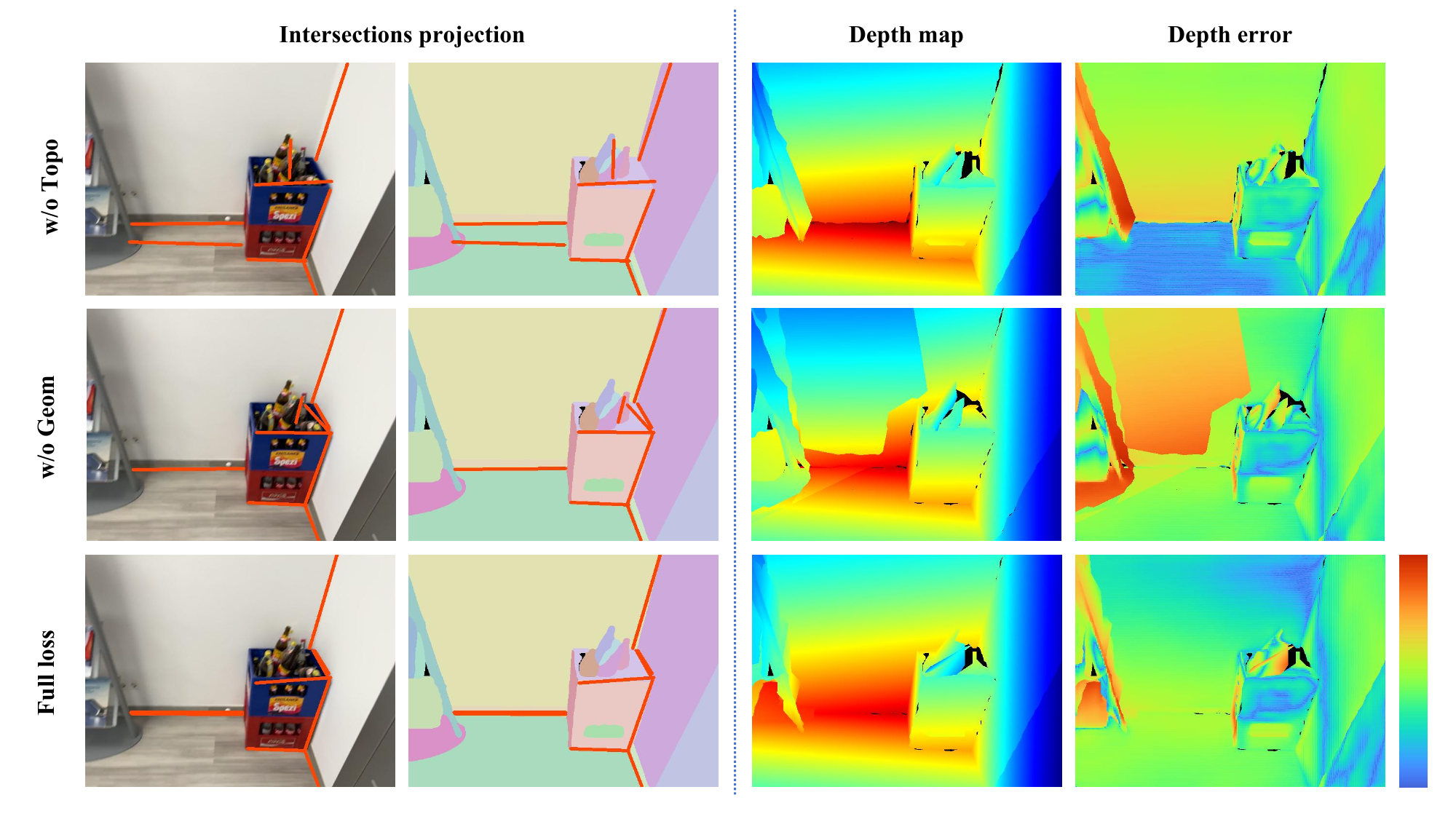}
    \caption{\textbf{Qualitative visualization of the ablation study.} The "w/o Topo" variant suffers from misaligned planar intersections and high residuals at structural boundaries. While "w/o Geom" improves intersection alignment, it fails to recover accurate absolute depth estimates due to the presence of several floating planes. Our full configuration synchronizes both constraints, achieving sharp, well-aligned boundaries and the lowest overall depth error.}
    \label{fig:ablation}
\vspace{-10pt}
\end{figure}

\begin{figure}[!ht]
    \centering
    \includegraphics[width=\linewidth]{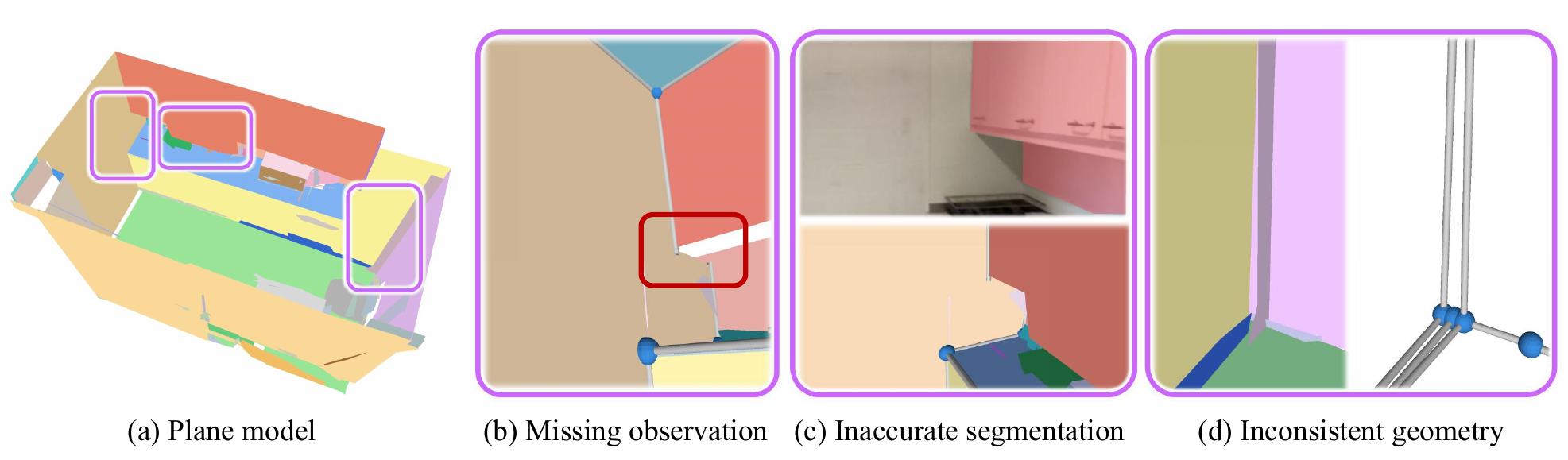} 
    \caption{\textbf{Limitation.} (a) Plane model generated by our method. (b) Incomplete connectivity results in non-watertight gaps at unobserved corners. (c) Inaccurate boundaries resulting from segmentation-induced errors during multi-view fusion. (d) Local inconsistencies generate multiple redundant intersection lines and vertices.}
    \label{fig:limitation}
    \vspace{-10pt}
\end{figure}

We conducted an ablation study to evaluate the individual contributions of the geometry loss and topology loss within our framework. The quantitative results are summarized in Table \ref{tab:ablation}.

The exclusion of topological constraints ("w/o Topo") leads to a measurable degradation in performance, with the Chamfer Distance (CD) increasing to $6.54$ and the F-score dropping to $64.98$. This validates our core insight: explicitly integrating these topological relationships as vital geometric constraints, rather than relegating them to optional post-processing, can effectively guide the 3DGS optimization landscape, enforce global structural coherence (which is evidenced by the improved topology score), and dramatically enhance robustness against noise. 
The "w/o Geom" variant yields the lowest F-score ($62.32$) and the poorest Planar Fidelity ($16.50$), underscoring the necessity of geometric regularization for anchoring planar positions.

Figure \ref{fig:ablation} visualizes the planar depth errors and the projection errors of plane-to-plane intersections for different variants. A key observation is that the "w/o Geom" configuration achieves deceptively low projection errors but exhibits relatively high 3D depth deviations, indicating that the model overfits to 2D image views without forming a coherent 3D structure. Conversely, while the "w/o Topo" variant maintains lower overall depth error than "w/o Geom," it exhibits significant projection inaccuracies at planar junctions. Notably, even "w/o Topo" depth appears plausible, our full configuration identifies and corrects its subtle geometric discrepancies. These results demonstrate that by jointly optimizing photometric, geometric, and topological consistency, TopoGS effectively synchronizes geometric precision with topological validity, resulting in a consistent high-fidelity scene structure.

\vspace{-10pt}
\subsection{Limitation}
It is worth noting that while our optimization encourages tight connectivity, the reconstructed models cannot be strictly watertight.
Multi-view images inherently contain occlusions and missing observations, especially at regions which have limited visibility.
Consequently, certain intersection lines or vertices may not be fully recovered, leading to non-watertight gaps at unobserved corners (see Fig.~\ref{fig:limitation} (b)).
One potential solution to this issue is to leverage geometry and shape priors learned from data to perform reconstruction and generation simultaneously.
Another common failure case is due to inaccurate segmentation or complicated local geometry, where we showed in Fig.~\ref{fig:limitation} (c) and (d).

\section{Conclusions}

In this paper, we present TopoGS, a novel framework for reconstructing structured planar models by integrating geometric and topological constraints within a 3D Gaussian Splatting optimization pipeline. Our method effectively addresses the limitations of existing two-stage and direct planar reconstruction approaches by jointly optimizing plane parameters and their topological relationships. Extensive experiments on the ScanNet++ demonstrate that TopoGS achieves state-of-the-art performance across multiple evaluation metrics, including geometry, planarity, segmentation, and topology. Future work will explore extending our framework to handle more complex scenes with non-planar surfaces and recovering missing regions from incomplete observations.

\section*{Acknowledgements}
This work was supported in part by Guangdong S\&T Program (2024B0101050004), ICFCRT (W2441020),  Guangdong Basic and Applied Basic Research Foundation (2023B1515120026), Shenzhen Science and Technology Program (KJZD2024 0903100022028, KQTD20210811090044003), and Guangdong Provincial Key Laboratory of Visual Media and Multidimensional Intelligence.


%
%
\bibliographystyle{splncs04}
\bibliography{main}

@String(PAMI  = {IEEE Trans. Pattern Anal. Mach. Intell.})

@String(CVPR  = {IEEE Conf. Comput. Vis. Pattern Recog.})

@String(ICCV  = {Int. Conf. Comput. Vis.})

@String(ECCV  = {Eur. Conf. Comput. Vis.})

@String(ICLR  = {Int. Conf. Learn. Represent.})

@String(BMVC  = {Brit. Mach. Vis. Conf.})

@String(TOG   = {ACM Trans. Graph.})

@String(TVCG  = {IEEE Trans. Vis. Comput. Graph.})

@String(CGF   = {Comput. Graph. Forum})

@inproceedings{xie2022planarrecon,
  title={Planarrecon: Real-time 3d plane detection and reconstruction from posed monocular videos},
  author={Xie, Yiming and Gadelha, Matheus and Yang, Fengting and Zhou, Xiaowei and Jiang, Huaizu},
  booktitle=CVPR,
  pages={6219--6228},
  year={2022}
}

@article{chen2024pgsr,
  title={Pgsr: Planar-based gaussian splatting for efficient and high-fidelity surface reconstruction},
  author={Chen, Danpeng and Li, Hai and Ye, Weicai and Wang, Yifan and Xie, Weijian and Zhai, Shangjin and Wang, Nan and Liu, Haomin and Bao, Hujun and Zhang, Guofeng},
  journal = TVCG,
  volume={31},
  number={9},
  pages={6100--6111},
  year={2024},
  publisher={IEEE}
}

@inproceedings{ruan2025indoorgs,
  title={Indoorgs: Geometric cues guided gaussian splatting for indoor scene reconstruction},
  author={Ruan, Cong and Wang, Yuesong and Guan, Tao and Zhang, Bin and Ju, Lili},
  booktitle= CVPR,
  pages={844--853},
  year={2025}
}

@article{jin2025planargs,
  title={PlanarGS: High-Fidelity Indoor 3D Gaussian Splatting Guided by Vision-Language Planar Priors},
  author={Jin, Xirui and Jin, Renbiao and Li, Boying and Zou, Danping and Yu, Wenxian},
  journal={arXiv preprint arXiv:2510.23930},
  year={2025}
}

@inproceedings{yang2024depth,
  title={Depth anything: Unleashing the power of large-scale unlabeled data},
  author={Yang, Lihe and Kang, Bingyi and Huang, Zilong and Xu, Xiaogang and Feng, Jiashi and Zhao, Hengshuang},
  booktitle=CVPR,
  pages={10371--10381},
  year={2024}
}

@article{yang2024depth2,
  title={Depth anything v2},
  author={Yang, Lihe and Kang, Bingyi and Huang, Zilong and Zhao, Zhen and Xu, Xiaogang and Feng, Jiashi and Zhao, Hengshuang},
  journal={Advances in Neural Information Processing Systems},
  volume={37},
  pages={21875--21911},
  year={2024}
}

@article{lin2025depth3,
  title={Depth anything 3: Recovering the visual space from any views},
  author={Lin, Haotong and Chen, Sili and Liew, Junhao and Chen, Donny Y and Li, Zhenyu and Shi, Guang and Feng, Jiashi and Kang, Bingyi},
  journal={arXiv preprint arXiv:2511.10647},
  year={2025}
}

@inproceedings{wang2025vggt,
  title={Vggt: Visual geometry grounded transformer},
  author={Wang, Jianyuan and Chen, Minghao and Karaev, Nikita and Vedaldi, Andrea and Rupprecht, Christian and Novotny, David},
  booktitle=CVPR,
  pages={5294--5306},
  year={2025}
}

@article{wang2025pi,
  title={$\pi^3$: Scalable Permutation-Equivariant Visual Geometry Learning},
  author={Wang, Yifan and Zhou, Jianjun and Zhu, Haoyi and Chang, Wenzheng and Zhou, Yang and Li, Zizun and Chen, Junyi and Pang, Jiangmiao and Shen, Chunhua and He, Tong},
  journal={arXiv e-prints},
  pages={arXiv--2507},
  year={2025}
}

@inproceedings{bleyer2011patchmatch,
  title={Patchmatch stereo-stereo matching with slanted support windows.},
  author={Bleyer, Michael and Rhemann, Christoph and Rother, Carsten},
  booktitle=BMVC,
  volume={11},
  number={2011},
  pages={1--11},
  year={2011}
}

@inproceedings{luo2016efficient,
  title={Efficient deep learning for stereo matching},
  author={Luo, Wenjie and Schwing, Alexander G and Urtasun, Raquel},
  booktitle=CVPR,
  pages={5695--5703},
  year={2016}
}

@inproceedings{yao2018mvsnet,
  title={Mvsnet: Depth inference for unstructured multi-view stereo},
  author={Yao, Yao and Luo, Zixin and Li, Shiwei and Fang, Tian and Quan, Long},
  booktitle=ECCV,
  pages={767--783},
  year={2018}
}

@inproceedings{sayed2022simplerecon,
  title={Simplerecon: 3d reconstruction without 3d convolutions},
  author={Sayed, Mohamed and Gibson, John and Watson, Jamie and Prisacariu, Victor and Firman, Michael and Godard, Cl{\'e}ment},
  booktitle=ECCV,
  pages={1--19},
  year={2022},
  organization={Springer}
}

@inproceedings{stier2023finerecon,
  title={Finerecon: Depth-aware feed-forward network for detailed 3d reconstruction},
  author={Stier, Noah and Ranjan, Anurag and Colburn, Alex and Yan, Yajie and Yang, Liang and Ma, Fangchang and Angles, Baptiste},
  booktitle=ICCV,
  pages={18423--18432},
  year={2023}
}

@inproceedings{schnabel2007efficient,
  title={Efficient RANSAC for point-cloud shape detection},
  author={Schnabel, Ruwen and Wahl, Roland and Klein, Reinhard},
  booktitle=CGF,
  volume={26},
  number={2},
  pages={214--226},
  year={2007},
  organization={Wiley Online Library}
}

@article{verdie2015lod,
  title={LOD generation for urban scenes},
  author={Verdie, Yannick and Lafarge, Florent and Alliez, Pierre},
  journal = TOG,
  year={2015},
  publisher={Association for Computing Machinery}
}

@article{bauchet2020kinetic,
  title={Kinetic shape reconstruction},
  author={Bauchet, Jean-Philippe and Lafarge, Florent},
  journal=TOG,
  volume={39},
  number={5},
  pages={1--14},
  year={2020},
  publisher={ACM New York, NY, USA}
}

@article{chen2022reconstructing,
  title={Reconstructing compact building models from point clouds using deep implicit fields},
  author={Chen, Zhaiyu and Ledoux, Hugo and Khademi, Seyran and Nan, Liangliang},
  journal={ISPRS Journal of Photogrammetry and Remote Sensing},
  volume={194},
  pages={58--73},
  year={2022},
  publisher={Elsevier}
}

@inproceedings{he2024windpoly,
  title={Windpoly: Polygonal mesh reconstruction via winding numbers},
  author={He, Xin and Lv, Chenlei and Huang, Pengdi and Huang, Hui},
  booktitle=ECCV,
  pages={294--311},
  year={2024},
  organization={Springer}
}

@inproceedings{sun2021neuralrecon,
  title={Neuralrecon: Real-time coherent 3d reconstruction from monocular video},
  author={Sun, Jiaming and Xie, Yiming and Chen, Linghao and Zhou, Xiaowei and Bao, Hujun},
  booktitle=CVPR,
  pages={15598--15607},
  year={2021}
}

@inproceedings{ye2025neuralplane,
  title={Neuralplane: Structured 3d reconstruction in planar primitives with neural fields},
  author={Ye, Hanqiao and Liu, Yuzhou and Liu, Yangdong and Shen, Shuhan},
  booktitle=ICLR,
  year={2025}
}

@inproceedings{zanjani2025planar,
  title={Planar gaussian splatting},
  author={Zanjani, Farhad G and Cai, Hong and Ackermann, Hanno and Mirvakhabova, Leila and Porikli, Fatih},
  booktitle={2025 IEEE/CVF Winter Conference on Applications of Computer Vision},
  pages={8905--8914},
  year={2025},
  organization={IEEE}
}

@inproceedings{tan2025planarsplatting,
  title={Planarsplatting: Accurate planar surface reconstruction in 3 minutes},
  author={Tan, Bin and Yu, Rui and Shen, Yujun and Xue, Nan},
  booktitle=CVPR,
  pages={1190--1199},
  year={2025}
}

@article{gan2025gsplane,
  title={GSPlane: Concise and Accurate Planar Reconstruction via Structured Representation},
  author={Gan, Ruitong and Peng, Junran and Liu, Yang and Luo, Chuanchen and Li, Qing and Zhang, Zhaoxiang},
  journal={arXiv preprint arXiv:2510.17095},
  year={2025}
}

@article{taktasheva20253d,
  title={3D Gaussian Flats: Hybrid 2D/3D Photometric Scene Reconstruction},
  author={Taktasheva, Maria and Goli, Lily and Fiorini, Alessandro and Li, Zhen and Rebain, Daniel and Tagliasacchi, Andrea},
  journal={arXiv preprint arXiv:2509.16423},
  year={2025}
}

@article{hu2024metric3d,
  title={Metric3d v2: A versatile monocular geometric foundation model for zero-shot metric depth and surface normal estimation},
  author={Hu, Mu and Yin, Wei and Zhang, Chi and Cai, Zhipeng and Long, Xiaoxiao and Chen, Hao and Wang, Kaixuan and Yu, Gang and Shen, Chunhua and Shen, Shaojie},
  journal=PAMI,
  volume={46},
  number={12},
  pages={10579--10596},
  year={2024},
  publisher={IEEE}
}

@inproceedings{kirillov2023segment,
  title={Segment anything},
  author={Kirillov, Alexander and Mintun, Eric and Ravi, Nikhila and Mao, Hanzi and Rolland, Chloe and Gustafson, Laura and Xiao, Tete and Whitehead, Spencer and Berg, Alexander C and Lo, Wan-Yen and others},
  booktitle=ICCV,
  pages={4015--4026},
  year={2023}
}

@inproceedings{liu2019planercnn,
  title={Planercnn: 3d plane detection and reconstruction from a single image},
  author={Liu, Chen and Kim, Kihwan and Gu, Jinwei and Furukawa, Yasutaka and Kautz, Jan},
  booktitle=CVPR,
  pages={4450--4459},
  year={2019}
}

@inproceedings{yeshwanth2023scannet++,
  title={Scannet++: A high-fidelity dataset of 3d indoor scenes},
  author={Yeshwanth, Chandan and Liu, Yueh-Cheng and Nie{\ss}ner, Matthias and Dai, Angela},
  booktitle=ICCV,
  pages={12--22},
  year={2023}
}

@inproceedings{watson2024airplanes,
  title={Airplanes: Accurate plane estimation via 3d-consistent embeddings},
  author={Watson, Jamie and Aleotti, Filippo and Sayed, Mohamed and Qureshi, Zawar and Mac Aodha, Oisin and Brostow, Gabriel and Firman, Michael and Vicente, Sara},
  booktitle=CVPR,
  pages={5270--5280},
  year={2024}
}

@article{kerbl20233dgs,
  title={3d gaussian splatting for real-time radiance field rendering.},
  author={Kerbl, Bernhard and Kopanas, Georgios and Leimk{\"u}hler, Thomas and Drettakis, George and others},
  journal=TOG,
  volume={42},
  number={4},
  pages={139--1},
  year={2023}
}

@article{meanshift,
  author={Comaniciu, D. and Meer, P.},
  journal=PAMI, 
  title={Mean shift: a robust approach toward feature space analysis}, 
  year={2002},
  volume={24},
  number={5},
  pages={603-619}
}
\end{document}